\UseRawInputEncoding
\pdfoutput=1

\documentclass[lettersize,journal]{IEEEtran}
\usepackage{amsmath,amsfonts}
\usepackage{algorithmic}
\usepackage{algorithm}
\usepackage{array}
\usepackage[caption=false,font=normalsize,labelfont=sf,textfont=sf]{subfig}
\usepackage{textcomp}
\usepackage{stfloats}
\usepackage{url}
\usepackage{verbatim}
\usepackage{graphicx}
\usepackage{cite}
\usepackage{color}
\usepackage{doi}
\usepackage{amsmath}
\begin{document}

\title{Global Relation Modeling and Refinement for Bottom-Up Human Pose Estimation}

\author{Ruoqi Yin, Jianqin Yin*,~\IEEEmembership{Member,~IEEE,}
\thanks{*Corresponding author.}
}



\maketitle

\begin{abstract}
In this paper, we concern on the bottom-up paradigm in multi-person pose estimation (MPPE). Most previous bottom-up methods try to consider the relation of instances to identify different body parts during the post processing, while ignoring to model the relation among instances or environment in the feature learning process. In addition, most existing works adopt the operations of upsampling and downsampling. During the sampling process, there will be a problem of misalignment with the source features, resulting in deviations in the keypoint features learned by the model. 

To overcome the above limitations, we propose a convolutional neural network for bottom-up human pose estimation. It invovles two basic modules: (i) Global Relation Modeling (GRM) module globally learns relation (e.g., environment context, instance interactive information) among region of image by fusing multiple stages features in the feature learning process. It combines with the spatial-channel attention mechanism, which focuses on achieving adaptability in spatial and channel dimensions. (ii) Multi-branch Feature Align (MFA) module aggregates features from multiple branches to align fused feature and obtain refined local keypoint representation. Our model has the ability to focus on different granularity from local to global regions, which significantly boosts the performance of the multi-person pose estimation. Our results on the COCO and CrowdPose datasets demonstrate that it is an efficient framework for multi-person pose estimation. 
\end{abstract}


\section{Introduction}
2D multi-person pose estimation aims at locating anatomical keypoints of all persons in natural images. As a fundamental task for human motion recognition, human-computer interaction, sign language recognition, and human behavior understanding \cite{ref1,ref2,ref3}, it receives increasing attention in recent years. With rich and longstanding studies, it makes enormous progress in pose estimation, benefiting from the more abundant datasets and the development of deep convolutional neural networks \cite{ref4,ref5,ref6,ref7}.

As one of the fundamental pipelines for 2D multi-person pose estimation, the bottom-up methods \cite{ref8,ref9,ref10,ref11} first detect all keypoint locations from the given multi-person image, and then group them to the corresponding human instance according to the instance relation. Since the bottom-up methods do not require an additional detection model, it is much more efficient than the top-down methods. Therefore, it attracts the attention of scholars in recent years. 

\begin{figure*}[!t]
\centering
\includegraphics[width=7in]{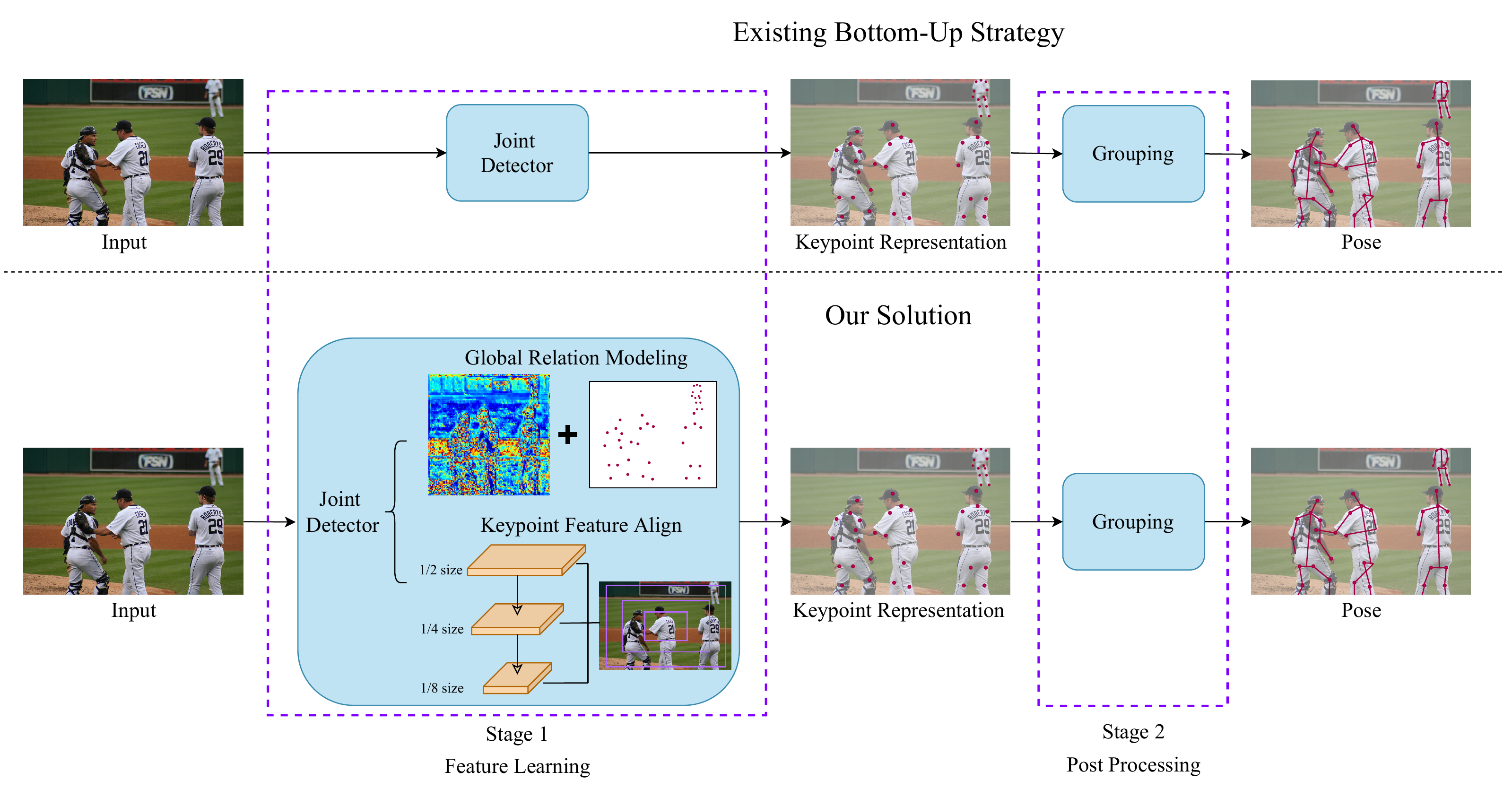}%
\caption{Top: the workflow of previous bottom-up methods; Bottom: our model focuses on modeling global relation and aligning multi-resolution fused feature.}
\label{fig1}
\end{figure*}

Although the existing methods achieve good performance, they ignore an important issue that modeling the global relation (e.g., environment context, instance interactive information) in the feature learning process. As shown in Fig. \ref{fig1}, the bottom-up paradigm is a two-stage process, which contains feature learning (stage 1) and post processing (stage 2). The previous bottom-up methods group the keypoints by learning part affinity fields \cite{ref8}, part association fields \cite{ref11}, and associative embedding \cite{ref9} in the post processing (stage 2). They ignore to model the relation among instances or environment in the keypoint feature learning process (stage 1). However, since the workflow of bottom-up methods make it difficult to locate each keypoint correctly for cluttered background and self occlusion, we argue that obtaining global relation during the feature learning process is important, the clues derived from the relation in the image (e.g., enviroment context, instance interactive information) can help to infer some invisible keypoints, and also group person parts to the correct instance. Thus, we propose a Global Relation Modeling module to learn global relation, in this way, pose estimation network can consider the relation among instances or environment in early stage, thereby improving the final performance.

Although the convolutional neural network (CNN) currently used for pose estimation achieve high performance, the extracted features have a strong bias on regional texture, resulting in a significant loss of contextual relation for visual objects \cite{ref12}. To address this limitation, previous works propose aggregation modules \cite{ref13,ref14,ref15}, but the simple implementation of self-attention in these approaches hinders the modeling of discriminative contextual relation. They rely on the output of the last layer of the network, ignoring the rich location information contained in the intermediate stages. In order to effectively integrate multi-stage spatial and semantic features and globally obtain relation among instances or environment, we first use a novel backbone network to generate feature maps at multiple stages, and then use the Global Relation Modeling module to collect these features to strengthen the representation of global relation. The feature maps at previous stages could be deemed as a prior to support prediction in the following stage. Clues derived from multi-stage features contain rich contextual relation, which can be used to infer the locations of invisible keypoints, and play an important role in fusing low-stage spatial signals and high-stage semantic information.

In addition, the current multi-person pose estimation methods concern on aggregating multi-scale features (e.g., HRNet \cite{ref4} and Stacked Hourglass network \cite{ref16}). Strong semantic information is obtained by downsampling, and high resolution recovery location information is recovered by upsampling, in this way, features of different levels are fully aggregated. However, there is a gap between the receptive fields of features from different resolutions, which may lead to problem with feature misalignment when fusing features. To alleviate this problem, we propose a Multi-branch Feature Align module, which aggregates features from multiple branches to align fused feature and obtain refined local keypoint representations. Combining the proposed two modules not only enhances local feature representation but simultaneously aggregates multi-stage feature for global perception, which is more compact and efficient.

\begin{figure*}[!t]
\centering
\includegraphics[width=7.2in]{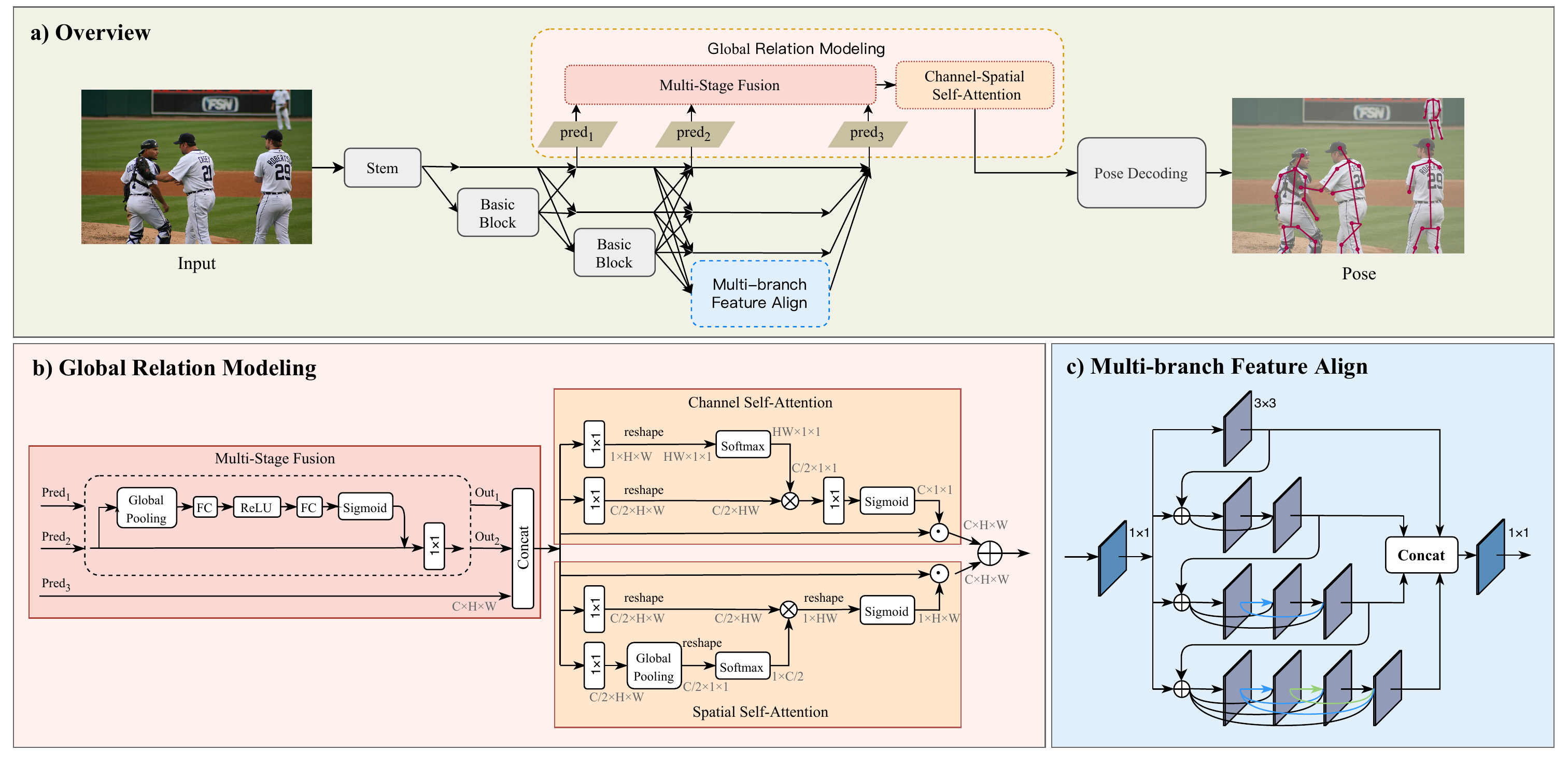}%
\caption{a) The overall proposed architecture. It invovles two basic modules: b) Global Relation Modeling (GRM) module globally learns relation (e.g., environment context, instance interactive information) among region of image by fusing multiple stages features in the feature learning process. It combines with the spatial-channel attention mechanism. c) Multi-branch Feature Align (MFA) module aggregates feature from multiple branches to align fused feature and obtain refined local keypoint representation.}
\label{fig2}
\end{figure*}

In a summary, we propose a convolutional neural network for bottom-up human pose estimation, as shown in Fig. \ref{fig2}, which mainly consists of three parts: (i) In order to mine the effectiveness of cross-stage feature interaction on pose estimation, we inherit the multi-stage feature extraction process from HRNet, which is different from the previous high-to-low or low-to-high process. It maintains high-resolution output throughout the whole process, integrates multi-scale information, and does not need intermediate supervision. In this way, features of different resolutions are fully aggregated. (ii) Global Relation Modeling module collects features from multiple stages to learn global relation in the feature learning process. This enables the model to not only capture keypoint locations information but also the contextual relation in the image simultaneously. It combines with the spatial-channel attention mechanism, which focuses on achieving adaptability in spatial and channel dimensions. (iii) Multi-branch Feature Align module aggregates feature from multiple branches to align fused feature and obtain refined local keypoint representations. We also conduct experiments to show that our method achieves good performance at accuracy. The main contributions of our work can be summarized as follows:

1) In our bottom-up approach, we design a light-weight backbone network, inheriting the multi-stage feature extraction process from HRNet, which aggregates features of different resolutions. 

2) We propose a Global Relation Modeling (GRM) module to aggregate features from multiple stages to obtain the guidance of global relation among instances or environment in the feature learning process. It combines with the spatial-channel attention mechanism.

3) We propose a Multi-branch Feature Align (MFA) module, which aggregates feature from multiple branches to align fused feature and obtain refined local keypoint representations. 

4) We demonstrate the effectiveness of our model on the challenging COCO and CrowdPose dataset. Our model outperforms all other bottom-up methods.

\section{Related Work}
{\bf{Bottom-up Human Pose Estimation}}. Bottom-up \cite{ref8,ref9,ref10,ref11} is one of the fundamental pipelines for multi-person pose estimation, which starts by detecting all keypoints from the input image simultaneously, followed by grouping them into individual instances. Since they do not rely on human detectors, the computing time is not affected by the number of persons, bottom-up methods may have more potential superiority on speed. But they have to tackle the complicated post processing problem.

The Convolutional Pose Machines (CPM) approach \cite{ref17} used CNN to extract feature and context information, and presented the prediction results in the form of heatmap. Based on the work \cite{ref17}, the OpenPose method \cite{ref8} combined the Part Affinity Fields (PAF), which can learn to associate body parts with the people in the image. Other methods further developed PAF, such as Pif-Paf \cite{ref11} and Associative Embedding \cite{ref9}. Recently, High-Resolution Network (HRNet) \cite{ref4}, a multi-scale feature representation approach for a larger Field-of-View, attracted extensive attention in the field of computer vision. For instance, HigherHRNet \cite{ref18} learned scale-aware representations using HRNet, and further used transposed convolution to generate higher-resolution features to improve accuracy. The recent Disentangled Keypoint Regression (DEKR) method \cite{ref19} used HRNet as the backbone to learn disentangled representations for each keypoint, and utilized adaptively activated pixels to directly regress the position of keypoints for each pixel in the image.

{\bf{Contextual Relation Aggregation}}. The contextual relation is generally referred to as regions surrounding the target locations, object-scene relations, and object-object interactions \cite{ref20,ref21,ref22,ref23,ref24}. In the literature, the contextual relation has been proved essential for vision tasks such as image classification \cite{ref25}, object detection \cite{ref26} and human pose estimation \cite{ref27}. However, previous works usually used manually designed multi-context representations, e.g., multiple bounding boxes \cite{ref27} or multiple image crops \cite{ref25}, and henced lack of flexibility and diversity for modeling the multi-context representations. Recently, many methods \cite{ref28,ref29,ref30} of human pose estimation modeled contextual relation by concatenating multi-scale features. Newell et al. \cite{ref28} proposed a U-shape convolutional neural network named Hourglass. The network sampled the input image to a small resolution, then added them to low-level features after upsampling, combining features of uniform dimensions.  Later works such as Chen et al. \cite{ref29} combined inter-level features using a RefineNet. Sun et al. \cite{ref30} set up four parallel sub-networks. The features of these four sub-networks aggregated with each other through high-to-low or low-to-high way. In this work, we adopt larger region to capture global spatial configurations of object, while smaller region to focus on the local part appearance. Additionally, we adopt visual attention mechanism to focus on regions which are image dependent and adaptiving for multi-context modeling.

\section{METHODOLOGY}
\subsection{Overview}
Multi-person pose estimation aims to predict the human poses, where each pose consists of $K$ keypoints, (e.g., shoulder, elbow, etc) from an image $\mathbf{I}$ of size $W\times H\times 3$. The proposed bottom-up method, illustrated in Fig. \ref{fig2}, is particularly effective for multi-person pose estimation. It inherits the multi-stage feature extraction process from HRNet to mine the effectiveness of cross-stage feature interaction on pose estimation. We propose two improvements: first, we propose a Global Relation Modeling (GRM) module to fuse features from multiple stages to obtain the guidance of global relation during the feature learning process. It also combines with the spatial-channel attention mechanism. Second, we propose a Multi-branch Feature Align (MFA) module, which aggregates feature from multiple branches to align fused feature and obtain refined local keypoint representations. Additionally, our approach integrates improvements in decoupling keypoint regression \cite{ref19}, which activates pixels in keypoint areas by adaptive convolution in a pixel-by-pixel space converter for person localization and parts association.

The  proposed architecture is shown in Fig. \ref{fig2}$(a)$. The input image $\mathbf{I}$ of size $W\times H\times 3$ is initially fed into a light-weight backbone network we design, which inherits the multi-stage feature extraction process from HRNet. The HRNet-based architecture mainly consists of $4$ stages (stem is regarded as stage $1$ here). In stage $n$, there are $n$ branches to process feature maps of different resolutions respectively. Let $\mathbf{A}_{n}
=[\mathbf{a}_{1}, \mathbf{a}_{2}, \dots, \mathbf{a}_{n}]$ denote the number of blocks for each branch in stage $n$ before fusion ($\mathbf{a}_{i}$ stands for the number of blocks in branch $i$). Then we can define the configuration of the whole multi-branch architecture as $\mathbf{A} =\{\mathbf{A}_{1}, \mathbf{A}_{2},  \mathbf{A}_{3}, \mathbf{A}_{4}\}$. In order to focus on the performance with lower computation, we take a shrinking process to make the whole network similar to a single-branch network \cite{ref37}, which only retains the convolution operation of the last branch in each stage. We list the original configuration $\mathbf{A}_{ori}$ and our configuration $\mathbf{A}_{our}$ in detail below:
\begin{align}
\begin{split}
	HRNet: \mathbf{A}_{ori} = \{[4], [4, 4], [4, 4, 4], [4, 4, 4, 4]\}\\
 Ours: \mathbf{A}_{our} = \{[4], [0, 4], [0, 0, 4], [0, 0, 0, 4]\}
\end{split}
\end{align}

Previous methods regress the heatmaps simply from the high-resolution representations output by the last stage. In order to mine global relation among instances or environment in the feature learning process, we use GRM module to converge features from multiple stages to learn global relation:
\begin{align}
	\mathbf{F}_{GRM} & = \operatorname{GRM}({p}_1, {p}_2, {p}_3).
\end{align}
where ${p}_{i}$ represents the highest resolution representation output by the stage $i+1 
 (i=1, 2, 3)$. $\operatorname{GRM}(\cdot)$ is GRM module.

Besides, the HRNet-based backbone has many upsampling and downsampling operations, which may cause the problem of feature misalignment. So we design MFA module to align fused features and obtain refined keypoint locations information. Concretely, in the block of the final stage ($4$th stage), it combines the rich features of multi-stage and multi-resolution, we replace the original Basic Block with MFA:
\begin{align}
	\mathbf{F}_{MFA} & = \operatorname{MFA}(\mathbf{x}^{4}_{4}).
\end{align}
where $\operatorname{MFA}(\cdot)$ is MFA module, $\mathbf{x}^{b}_{s}$ is the fusion feature before refinement.($s=4$ denotes the $4$th stage, $b=4$ denotes the $4$th block).

Finally, the network extracts keypoint features processed by decoders of disentagled keypoint regression, and adds a $1 \times 1$ convolution to predict heatmaps and tagmaps similar to \cite{ref19}. We follow \cite{ref9} to use associative embedding for keypoint grouping. The grouping process clusters identity-free keypoints into individuals by grouping keypoints whose tags have small ${l}_{2}$ distance. The loss function follows the setting of \cite{ref19}.

\subsection{Global Relation Modeling (GRM)}
The bottom-up framework identifies different human instances during the post processing, while ignoring to model the global relation among instances or environment in the feature learning process. To alleviate this problem, we propose a Global Relation Modeling (GRM) module to aggregate features from multiple stages to obtain the guidance of global relation in the feature learning process. It combines with the spatial-channel attention mechanism.

The Global Relation Modeling (GRM) module is shown in Fig. \ref{fig2}$(b)$. Two separate information flows are introduced from the backbone network in the stage 2 and stage 3 highest resolution feature maps, going through the SE Block \cite{ref31} and conv$3\times3$ operations, respectively, for rich contextual relation. We add the feature maps processed in the first two stages as prior information and concat with the feature map in the last stage:
\begin{align}
	\mathbf{F}_{all} & = \operatorname{Concat}((\mathbf{W}_{a}({F}_{SE}({p}_1))+\mathbf{W}_{b}({F}_{SE}({p}_2))), {p}_3).
\end{align}
where $\mathbf{W}_{a}$ and $\mathbf{W}_{b}$ are $3\times3$ convolution layers, ${F}_{SE}(\cdot)$ is SE Block, the function ${F}_{SE}(\cdot)$ is implemented as: ${p}_{i} \rightarrow {GAP} \rightarrow {RELU} \rightarrow {FC} \rightarrow {Sigmoid} \rightarrow {+ {p}_{i}} \rightarrow Conv1\times 1$, ${p}_{i}$ is the highest resolution output feature from stage $i+1 (i=1, 2, 3)$.

Then the feature map goes through a dual attention module. The proposed channel-spatial attention module processes different levels of feature maps from the backbone through two branches including channel self-attention and spatial self-attention. Low-level and high-level features are represented at the same resolution, fusing for more comprehensive global relation helps achieve a refined result for keypoint localization. By extracting the global relation around each detected joint and strengthening the relation between the associated joints, the network can better infer the confidence map and offset map of individual keypoints during subsequent adaptive regression. The proposed channel-spatial attention mechanism is also shown in Fig. \ref{fig2}$(b)$. And the details are as following:

{\bf{Channel-only branch}}
\begin{align}
	\mathbf{A}_{ch}(\mathbf{X}) & = {F}_{SG}[\mathbf{W}_{c}({r}_1(\mathbf{W}_{d}(\mathbf{X}))) \times [{F}_{SM}({r}_2(\mathbf{W}_{e}(\mathbf{X})))].
\end{align}

Where $\mathbf{W}_{c}$, $\mathbf{W}_{d}$ and We are $1\times1$ convolution layers, the internal number of channels in $\mathbf{W}$ is $C/2$. ${r}_1$ and ${r}_2$ are two tensor reshape operators, and “$\times$” is the matrix dot-product operation. ${F}_{SG}(\cdot)$ is a Sigmiod operator and ${F}_{SM}(\cdot)$ is a SoftMax operator: ${F}_{SM}(\mathbf{X})= \sum_{j=1}^{N_p}{\frac{e^{x_j}}{\sum_{m=1}^{N_p}{e^{x_m}}}{x_j}}$.

\hspace*{\fill}

{\bf{Spatial-only branch}}
\begin{align}
	\mathbf{A}_{sp}(\mathbf{X}) & = {F}_{SG}[{r}_3({F}_{SM}({r}_1({F}_{GP}(\mathbf{W}_{e}(\mathbf{X})))) \times {r}_2(\mathbf{W}_{d}(\mathbf{X}))].
\end{align}

Where $\mathbf{W}_{d}$ and $\mathbf{W}_{e}$ are standard $1\times1$ convolution layers. ${r}_1$, ${r}_2$ and ${r}_3$ are three tensor reshape operators, and “$\times$” is the matrix dot-product operation. ${F}_{SG}(\cdot)$ is a Sigmiod operator, ${F}_{SM}(\cdot)$ is a SoftMax operator, and ${F}_{GP}(\cdot)$ is a global pooling operator: $ F_{GP}(\mathbf{X})= \frac{1}{H\times W} \sum_{i=1}^{H} \sum_{j=1}^{W} X(:,i,j)$.

\hspace*{\fill}

{\bf{Composition}}

The output of above two branches are composed under the parallel layout:
\begin{align}
	\mathbf{F}_{GRM} & = \mathbf{A}_{ch}(\mathbf{X})\odot\mathbf{X} + \mathbf{A}_{sp}(\mathbf{X})\odot\mathbf{X}.
\end{align}

Where $\odot$ is a multiplication operator, and “+” is the element-wise addition operator.

We use GRM to aggregate features from multiple stages to obtain the guidance of global relation, this strategy effectively propagate rich information from the early stages to the current stage, achieve adaptability in spatial and channel dimensions. In the deeper stages, the backbone network also keeps strong semantic features, which are aggregated with rich spatial information features in the early stages, to enable multi-level feature extraction. Besides capturing multi-level features in encoder, introduces an efficient attention mechanism and achieves adaptability in spatial and channel dimensions. This enables the model to not only capture keypoints location information but also the contextual relation in the image simultaneously.

\subsection{Multi-branch Feature Align (MFA)}
Densely connected networks (DCN) are proposed in \cite{ref32}. In a DCN, feature reuse is achieved through the connection of features on channel, which establishes a dense connection of all the previous layers to the latter. It is found that the dense network has better performance than the normal network, which not only avoids the vanishing gradient problem, but also improves the parameter efficiency of the network \cite{ref32}. In our model, we propose a Multi-branch Feature Align (MFA) module that is used in the last stage of the backbone. The MFA is designed for aligning feature and learning refined local representations by feature reuse. 

An illustrative diagram of the proposed MFA module is shown in Fig. \ref{fig2}$(c)$. We design this module mainly inspired by this work \cite{ref32} and \cite{ref33}. The Dense Step Block firstly puts the feature through a conv$1\times1$ (convolutional layer with kernel size $1\times1$) to fuse pixel-level features of different scales to reduce feature perturbation. Then, it divides the feature into $4$ branches on channel dimension. On the $n$th branch, we use $n$ layers conv$3\times3$ to generate the feature map, these $3\times3$ convolutions are designed in the form of dense convolution. Setting up different numbers of convolution operations in different branches to obtain varying degrees of detail, including low-level pose information and high-level semantic information. The ability to express features of the same level is effectively enhanced by dense convolution and cross-layer connections. 

Consequently, Let ${x}^{l}_{b}$ be the input of the $l$th layer in the $b$th branch. The $1$st layer in each branch ${x}^{1}_{b}$ receives the original input ${x}_0$ and the output of the $(b-1)$th branch:
\begin{align}
	{x}^{1}_{b} & = {x}_0 + {o}_{b-1}.
\end{align}
And the other layers receive the feature-maps of all preceding layers, ${x}_{0},\dots,{x}^{l-1}_{b}$, as input on the $b$th branch:
\begin{align}
	{x}^{l}_{b} & = {F}_{DB}([{x}_0,\dots,{x}^{l-1}_{b}]).
\end{align}
where ${F}_{DB}$ refers DenseBlock, $[{x}_0, \dots, {x}^{l-1}_{b}]$ refers to the concatenation of the feature-maps produced in layers $0, \dots, l-1(l=2,3,4)$.

Benefit from the dense connection structure, small-gap receptive fields of features are fully fused resulting in delicate local representations. The output features are then concatenated to go through a conv$1\times1$.
\begin{align}
	\mathbf{F}_{MFA} & = \mathbf{W}(\operatorname{Concat}({o}_{1}, {o}_{2}, {o}_{3}, {o}_{4})).
\end{align}
where $\mathbf{W}$ is $1\times1$ convolution, ${o}_{b}(b=1,2,3,4)$ is the output of the $b$th branch.

Additionally, during the training process, the deeply connected structure contributes sufficient gradients, which benefits the keypoint localization task.

\begin{figure*}[t]
	\centering
	\includegraphics[height = 0.21\textwidth]{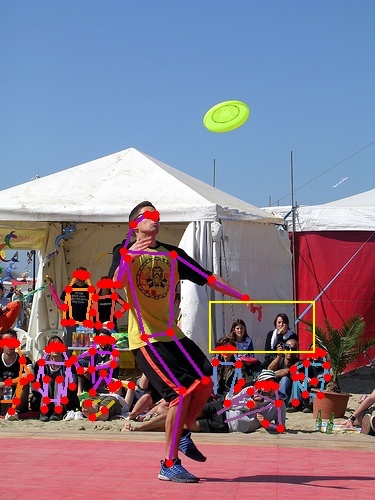}
	\includegraphics[height = 0.21\textwidth]{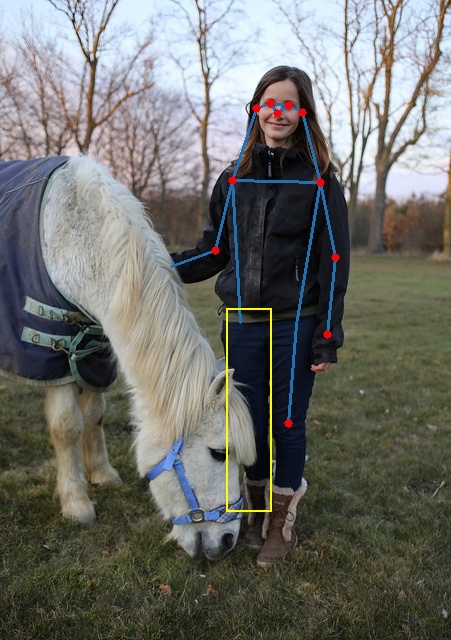}
    \includegraphics[height = 0.21\textwidth]{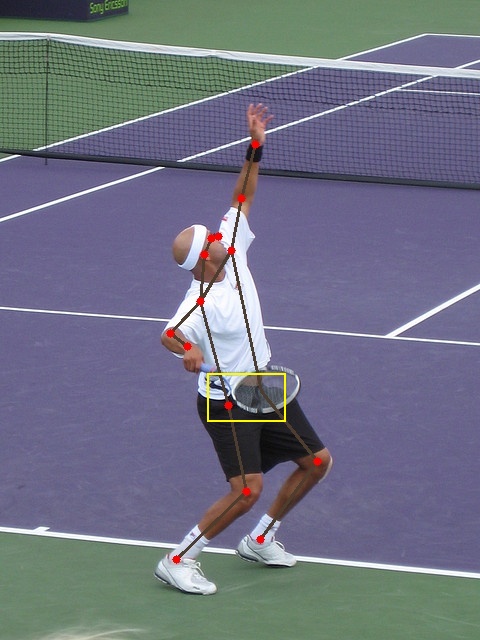}
    \includegraphics[height = 0.21\textwidth]{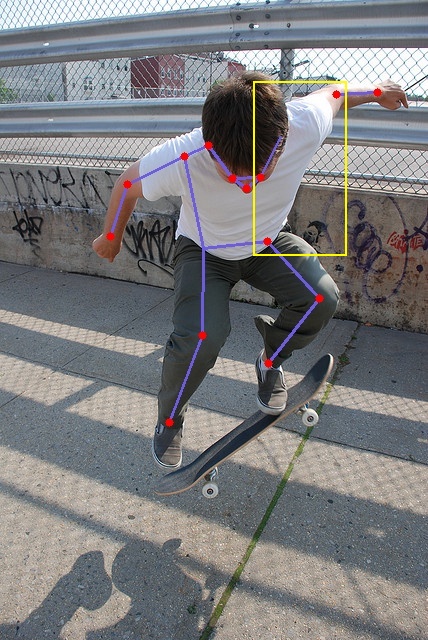}\\
    \includegraphics[height = 0.21\textwidth]{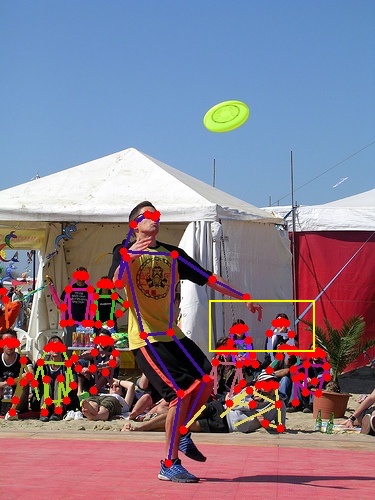}
	\includegraphics[height = 0.21\textwidth]{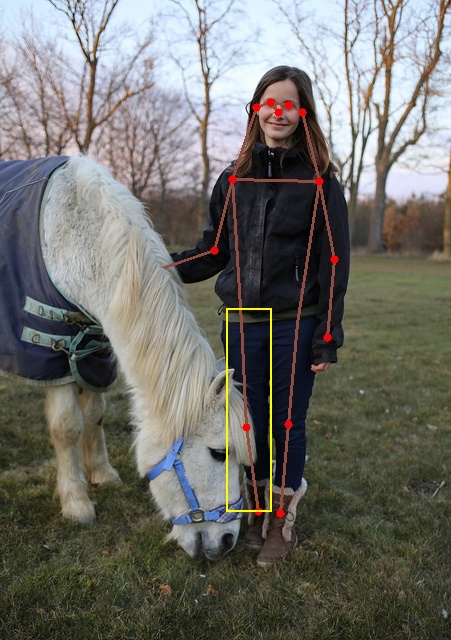}
    \includegraphics[height = 0.21\textwidth]{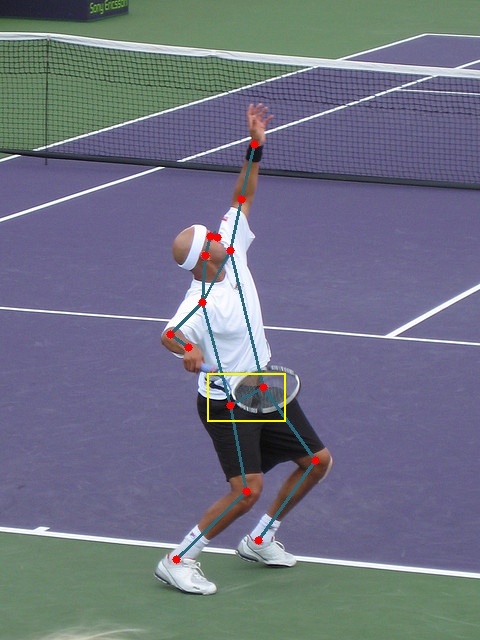}
    \includegraphics[height = 0.21\textwidth]{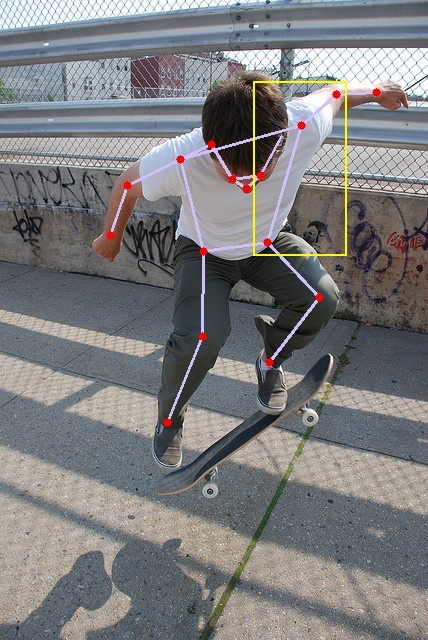}\\
	\includegraphics[height = 0.149\textwidth]{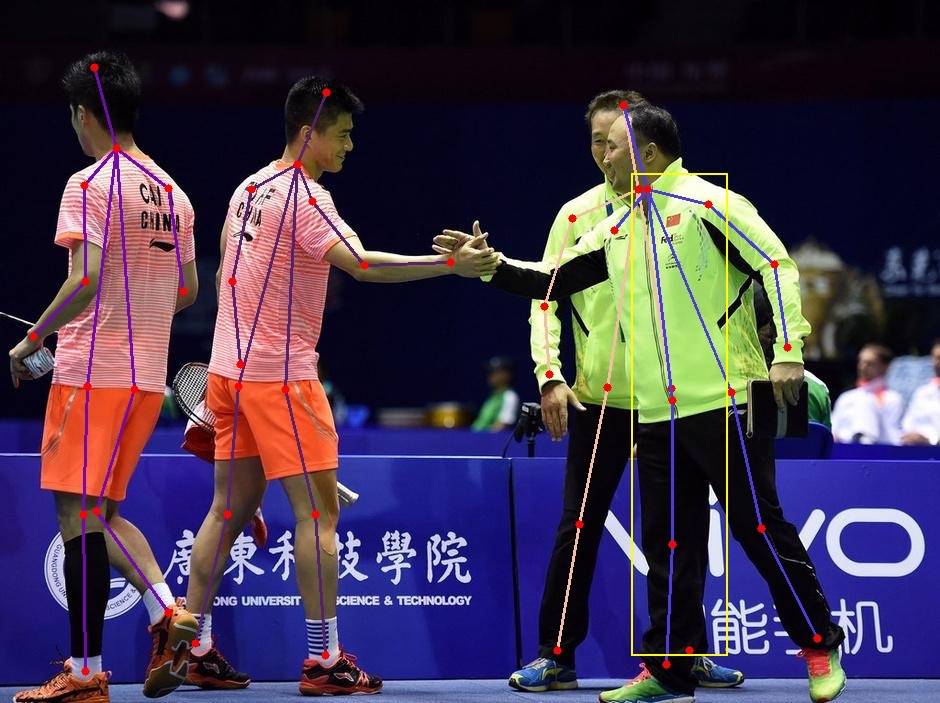}
	\includegraphics[height = 0.149\textwidth]{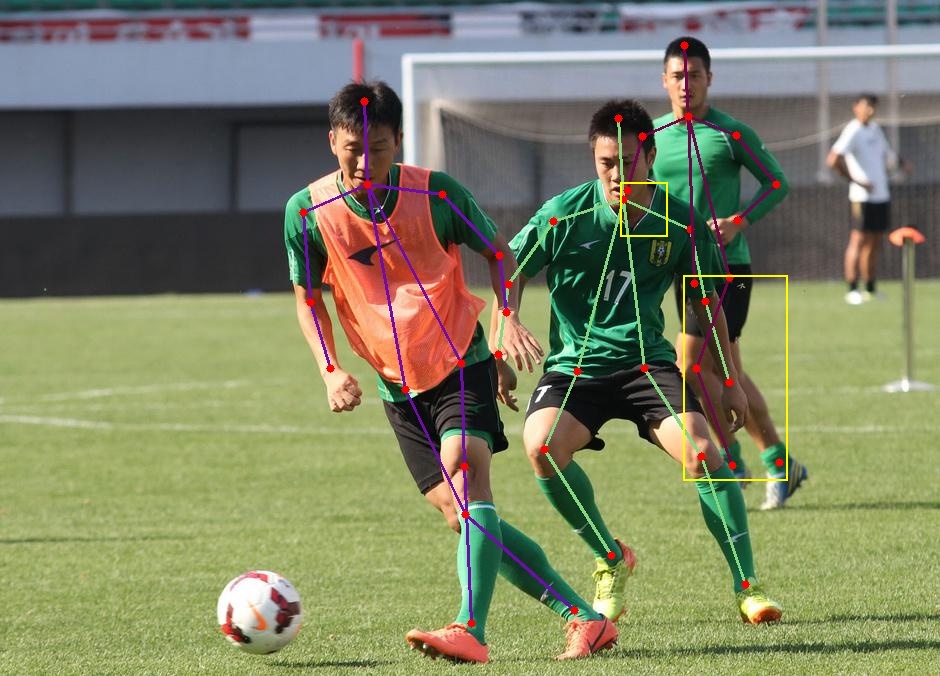}
    \includegraphics[height = 0.149\textwidth]{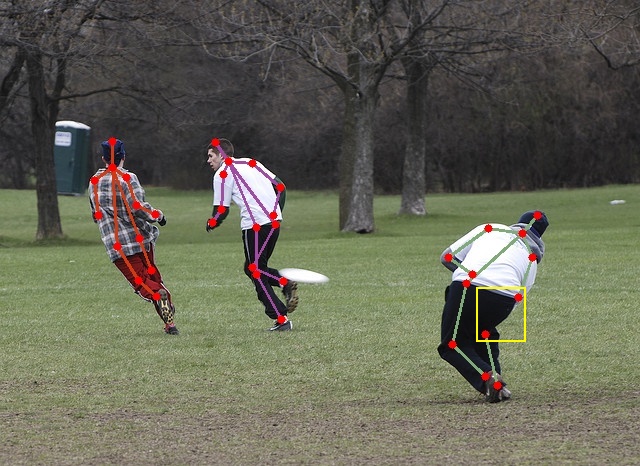}\\
    \includegraphics[height = 0.149\textwidth]{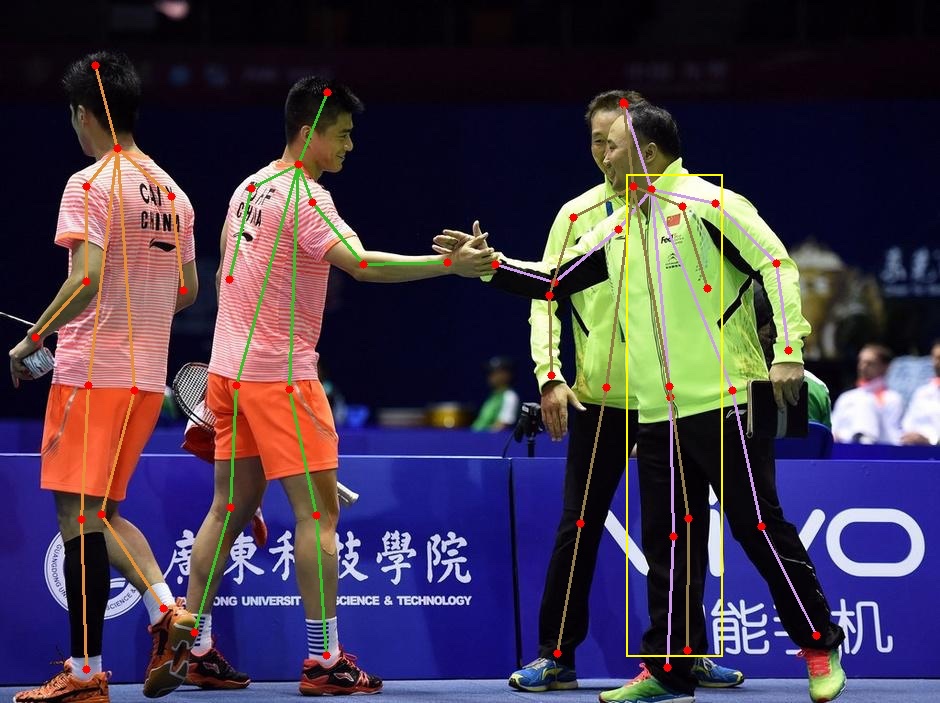}
	\includegraphics[height = 0.149\textwidth]{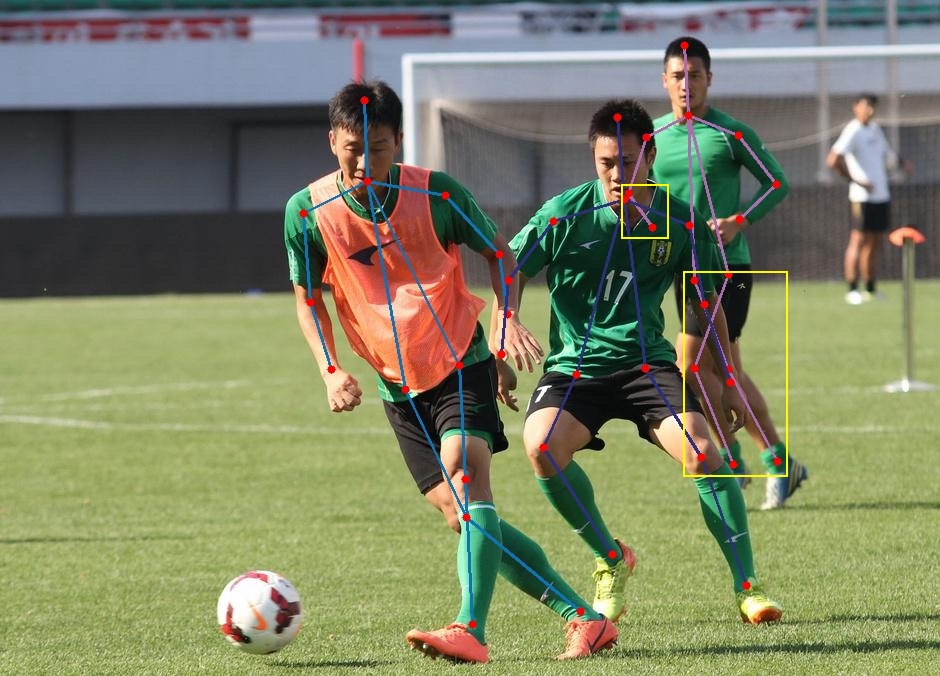}
    \includegraphics[height = 0.149\textwidth]{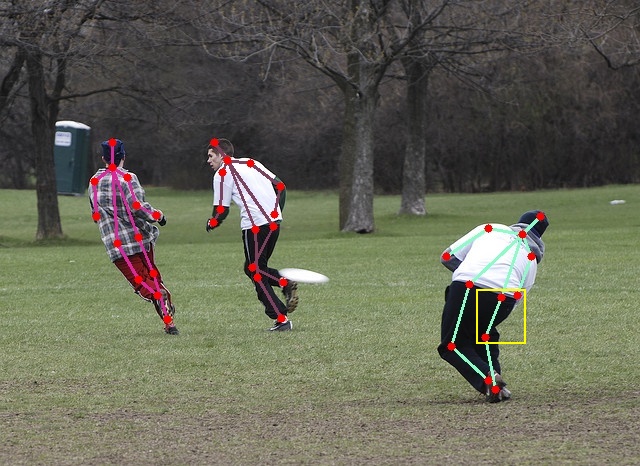}\\
	\caption{Qualitative results. Row1: Visualization results of the model \cite{ref19} on the COCO dataset. Row2: Visualization results of our method on the COCO dataset. Row3: Visualization results of the model \cite{ref19} on the CrowdPose dataset. Row4: Visualization results of our method on the CrowdPose dataset. Our model makes some occluded keypoints  be recognized very well, and the original wrong points be correctly divided into groups.}
    \label{fig3}
\end{figure*}

\section{EXPERIMENTS}
\subsection{Implementation Details}
\subsubsection{Dataset}{\bf{COCO.}} The COCO dataset \cite{ref34} is a popular multi-person pose estimation benchmark containing more than $200,000$ images and $250,000$ person instances labeling with $17$ keypoints. The COCO dataset is divided into three parts: the train set includes $57K$ images, the val set contains $5K$ images, and the test-dev set consists of $20K$ images. However, it contains rarely crowded scenarios, which we use to validate the generalization of our methods in unobstructed scenarios.

{\bf{CrowdPose.}} The CrowdPose dataset \cite{ref35} is a challenging benchmark with the goal to evaluate the robustness of methods in crowded scenes. The CrowdPose dataset contains $20K$ images and $80K$ persons labeled with $14$ keypoints. It contains $10K$, $2K$ and $8K$ images for train, val and test set. Following \cite{ref19}, we train our model on the train and validation splits combined, and report the final performance on the test set. 

\subsubsection{Evaluation metrics}
We use the standard evaluation metric Object Keypoint Similarity (OKS) to evaluate our models, which is defined as:
\begin{align}
\operatorname{OKS} = \frac{\sum_{i}\exp(-d_i^2/2s^2k_i^2)\delta(v_i > 0)}{\sum_i \delta(v_i > 0)}.
\end{align}
Here $d_i$ is the Euclidean distance between the detected keypoint and the corresponding ground truth, $v_i$ is the visibility flag of the ground truth, $s$ is the object scale, and $k_i$ is a per-keypoint constant that controls falloff. We report average precision and average recall scores with different thresholds and different object sizes: $\operatorname{AP}$, $\operatorname{AP}^{50}$, $\operatorname{AP}^{75}$, $\operatorname{AP}^{M}$, $\operatorname{AP}^{L}$, $\operatorname{AR}$, $\operatorname{AR}^{50}$, $\operatorname{AR}^{75}$, $\operatorname{AR}^{M}$ and $\operatorname{AP}^{L}$. The CrowdPose dataset is split into three crowding levels: easy, medium, hard, which stands for $\operatorname{AP}$ scores over easy, medium and hard instances, according to dataset annotations.

\subsubsection{Training details}
The data augmentation follows \cite{ref9} and includes random rotation $([30°, 30°])$, random scale $([0.75, 1.5])$ and random translation $([40, 40])$. We conduct image cropping to $512 \times 512$ for HRNet-W$32$ or $640 \times 640$ for HRNet-W$48$ with random flipping as training samples. HRNet-W$32$ and HRNet-W$48$, where $32$ and $48$ represent the widths of the high-resolution subnetworks in last three stages, respectively.

For COCO dataset, we use the Adam optimizer \cite{ref36}. The base learning rate is set as $1e-3$, and is dropped to $1e-4$ and $1e-5$ at the $90th$ and $120th$ epochs, respectively. The training process is terminated within $140$ epochs. For CrowdPose dataset, we use the Adam optimizer \cite{ref36}. The base learning rate is set as $1e-3$, and is dropped to $1e-4$ and $1e-5$ at the $200$th and $260$th epochs, respectively. The training process is terminated within $300$ epochs.

\subsubsection{Testing details}
We resize the short side of the images to $512/640$ and keep the aspect ratio between height and width, and compute the heatmap and pose positions by averaging the heatmaps and pixel-wise keypoint regressions of the original and flipped images. Following \cite{ref38}, we adopt three scales $0.5$, $1$ and $2$ in multi-scale testing. We average the three heatmaps over three scales and collect the regressed results from the three scales as the candidates.

\renewcommand{\arraystretch}{1.15}
	\begin{table*}[t]
		\centering
		\caption{Ablation Study on CrowdPose test dataset. ‘GRM’ represents the Global Relation Modeling module, ‘MFA’ indicates the Multi-branch Feature Align module.}
\setlength{\tabcolsep}{6.24pt}
		\label{tab1}
		\footnotesize
		\begin{tabular}{ c  c | c  c  c  c  c  c | c  c  c }
			\hline
			GRM & MFA & $\operatorname{AP}$ & $\operatorname{AP}^{50}$ & $\operatorname{AP}^{75}$ & $\operatorname{AP}^{E}$ & $\operatorname{AP}^{M}$ &
            $\operatorname{AP}^{H}$ &
            $\operatorname{AR}$ & 
            $\operatorname{AR}^{50}$ & 
            $\operatorname{AR}^{75}$\\
				\hline
				& &  $66.0$ & $86.2$ & $71.0$ & $73.2$ & $66.6$ & $57.7$ & $72.7$ & $91.1$ & $77.3$ \\ 
				\checkmark & &  $66.0$ & $86.3_{\textcolor{red}{\uparrow 0.1}}$ & $70.9$ & $73.3_{\textcolor{red}{\uparrow 0.1}}$ & $66.9_{\textcolor{red}{\uparrow 0.3}}$ & $57.6$ & $73.2_{\textcolor{red}{\uparrow 0.5}}$ & $92.0_{\textcolor{red}{\uparrow 0.9}}$ & $77.9_{\textcolor{red}{\uparrow 0.6}}$ \\ 
				& \checkmark &  $66.8_{\textcolor{red}{\uparrow 0.8}}$ & $86.6_{\textcolor{red}{\uparrow 0.4}}$ & $71.9_{\textcolor{red}{\uparrow 0.9}}$ & $74.0_{\textcolor{red}{\uparrow 0.8}}$ & $67.4_{\textcolor{red}{\uparrow 0.8}}$ & $58.7_{\textcolor{red}{\uparrow 1.0}}$ & $73.5_{\textcolor{red}{\uparrow 0.8}}$ & $91.6_{\textcolor{red}{\uparrow 0.5}}$ & $78.1_{\textcolor{red}{\uparrow 0.8}}$ \\
				\checkmark & \checkmark  & $66.8_{\textcolor{red}{\uparrow 0.8}}$ & $86.4_{\textcolor{red}{\uparrow 0.2}}$ & $71.9_{\textcolor{red}{\uparrow 0.9}}$ & $73.9_{\textcolor{red}{\uparrow 0.7}}$ & $67.5_{\textcolor{red}{\uparrow 0.9}}$ & $58.5_{\textcolor{red}{\uparrow 0.8}}$ & $73.9_{\textcolor{red}{\uparrow 1.2}}$ & $91.9_{\textcolor{red}{\uparrow 0.8}}$ & $78.6_{\textcolor{red}{\uparrow 1.3}}$ \\
				\hline 
			\end{tabular}
		\vspace{-.3cm}

	\end{table*}

\subsection{Results and Analyses}
\subsubsection{Ablation Study}

\renewcommand{\arraystretch}{1.15}
	\begin{table*}[t]
		\centering
		\caption{Comparisons on the CrowdPose test set.}
\setlength{\tabcolsep}{9.74pt}
		\label{tab2}
		\footnotesize
		\begin{tabular}{l|c|llllll|lll}
		    \hline
			Method & Input size &$\operatorname{AP}$ & $\operatorname{AP}^{50}$ & $\operatorname{AP}^{75}$ & $\operatorname{AP}^{E}$ & $\operatorname{AP}^{M}$ & $\operatorname{AP}^{H}$ & $\operatorname{AR}$ & $\operatorname{AR}^{50}$ & $\operatorname{AR}^{75}$\\
			\hline
			\multicolumn{10}{c}{single-scale testing}\\
			\hline
			OpenPose~\cite{ref8} & $-$ & $-$ & $-$ & $-$ & $62.7$ & $48.7$ & $32.3$ & $-$ & $-$ & $-$\\
			HrHRNet-W$48$~\cite{ref18} & $640$ & $65.9$ & $86.4$ & $70.6$ & $73.3$ & $66.5$ & $57.9$ & $-$ & $-$ & $-$\\
            DEKR-W$32$~\cite{ref19} & $512$ & $65.7$ & $85.7$ & $70.4$ & $73.0$ & $66.4$ & $57.5$ & $-$ & $-$ & $-$\\
            DEKR-W$48$ & $640$ & $67.3$ & $86.4$ & $72.2$ & $74.6$ & $68.1$ & $58.7$ & $-$ & $-$ & $-$\\
			\hline 
			Our approach (HRNet-W$32$) & $512$ & $\bf{66.8}$ & $\bf{86.4}$ & $\bf{71.9}$ & $\bf{73.9}$ & $\bf{67.5}$ & $\bf{58.5}$ & $\bf{73.9}$ & $\bf{91.9}$ & $\bf{78.6}$\\
			Our approach (HRNet-W$48$) & $640$ & $\bf{68.6}$ & $\bf{87.2}$ & $\bf{73.9}$ & $\bf{75.5}$ & $\bf{69.4}$ & $\bf{60.4}$ & $\bf{76.0}$ & $\bf{93.0}$ & $\bf{80.9}$\\
			\hline
			\multicolumn{10}{c}{multi-scale testing}\\
			\hline
			HrHRNet-W$48$ & $640$ & $67.6$ & $\bf{87.4}$ & $72.6$ & $75.8$ & $68.1$ & $58.9$ & $-$ & $-$ & $-$\\
            DEKR-W$32$ & $512$ & $67.0$ & $85.4$ & $72.4$ & $75.5$ & $68.0$ & $56.9$ & $-$ & $-$ & $-$\\
            DEKR-W$48$ & $640$ & $68.0$ & $85.5$ & $73.4$ & $76.6$ & $68.8$ & $58.4$ & $-$ & $-$ & $-$\\
			\hline
			Our approach (HRNet-W$32$) & $512$ & $\bf{68.2}$ & $\bf{85.9}$ & $\bf{73.7}$ & $\bf{76.5}$ & $\bf{69.2}$ & $\bf{58.2}$ & $\bf{75.5}$ & $\bf{92.1}$ & $\bf{80.5}$\\
			Our approach (HRNet-W$48$) & $640$ & $\bf{69.1}$ & $86.1$ & $\bf{74.6}$ & $\bf{77.0}$ & $\bf{70.2}$ & $\bf{59.5}$ & $\bf{76.7}$ & $\bf{92.4}$ & $\bf{81.7}$\\
			\hline
		\end{tabular}
		\vspace{-.3cm}
	\end{table*}

\renewcommand{\arraystretch}{1.15}
	\begin{table*}[t]
		\centering
		\caption{Comparisons on the COCO validation set.}
\setlength{\tabcolsep}{8.24pt}
		\label{tab3}
		\footnotesize
		\begin{tabular}{l|c|lllll|lllll}
		    \hline
			Method & Input size &$\operatorname{AP}$ & $\operatorname{AP}^{50}$ & $\operatorname{AP}^{75}$ & $\operatorname{AP}^{M}$ & $\operatorname{AP}^{L}$ & $\operatorname{AR}$ & $\operatorname{AR}^{50}$ & $\operatorname{AR}^{75}$ & $\operatorname{AR}^{M}$ & $\operatorname{AR}^{L}$\\
			\hline
			\multicolumn{10}{c}{single-scale testing}\\
			\hline
			PifPaf~\cite{ref11} & $-$ & $\bf{67.4}$ & $-$ & $-$ & $\bf{62.4}$ & $72.9$ & $-$ & $-$ & $-$ & $-$ & $-$\\
			HGG~\cite{ref39} & $512$ & $60.4$ & $83.0$ & $66.2$ & $-$ & $-$ & $64.8$ & $-$ & $-$ & $-$ & $-$\\
            PersonLab~\cite{ref10} & $1401$ & $66.5$ & $86.2$ & $71.9$ & $62.3$ & $73.2$ & $70.7$ & $-$ & $-$ & $65.6$ & $77.9$\\
            HrHRNet-W$32$ & $512$ & $67.1$ & $86.2$ & $73.0$ & $-$ & $-$ & $-$ & $-$ & $-$ & $61.5$ & $76.1$\\
            HrHRNet-W$48$ & $640$ & $69.9$ & $87.2$ & $76.1$ & $-$ & $-$ & $-$ & $-$ & $-$ & $65.4$ & $76.4$\\
			\hline 
			Our approach (HRNet-W$32$) & $512$ & $67.3$ & $\bf{86.4}$ & $\bf{74.4}$ & $67.1$ & $\bf{76.9}$ & $\bf{72.6}$ & $\bf{89.9}$ & $\bf{78.4}$ & $\bf{65.9}$ & $\bf{82.2}$\\
			Our approach (HRNet-W$48$) & $640$ & $\bf{70.1}$ & $\bf{88.0}$ & $\bf{76.5}$ & $\bf{65.9}$ & $\bf{77.8}$ & $\bf{75.5}$ & $\bf{91.6}$ & $\bf{81.1}$ & $\bf{70.2}$ & $\bf{83.5}$\\
			\hline
			\multicolumn{10}{c}{multi-scale testing}\\
			\hline
			HGG & $512$ & $68.3$ & $86.7$ & $75.8$ & $-$ & $-$ & $72.0$ & $-$ & $-$ & $-$ & $-$\\
            Point-Set~\cite{ref40} & $640$ & $69.8$ & $\bf{88.8}$ & $76.3$ & $\bf{65.9}$ & $76.6$ & $\bf{75.6}$ & $-$ & $-$ & $\bf{70.6}$ & $\bf{83.1}$\\
            HrHRNet-W$32$ & $512$ & $69.9$ & $87.1$ & $76.0$ & $-$ & $-$ & $-$ & $-$ & $-$ & $65.3$ & $77.0$\\
            HrHRNet-W$48$ & $640$ & $\bf{72.1}$ & $\bf{88.4}$ & $\bf{78.2}$ & $-$ & $-$ & $-$ & $-$ & $-$ & $67.8$ & $78.3$\\
			\hline
			Our approach (HRNet-W$32$) & $512$ & $\bf{69.8}$ & $86.8$ & $\bf{76.8}$ & $65.4$ & $\bf{77.1}$ & $75.2$ & $\bf{91.0}$ & $\bf{81.1}$ & $69.8$ & $\bf{83.1}$\\
			Our approach (HRNet-W$48$) & $640$ & $71.2$ & $87.5$ & $77.7$ & $\bf{67.6}$ & $\bf{77.8}$ & $\bf{77.0}$ & $\bf{92.1}$ & $\bf{82.6}$ & $\bf{72.1}$ & $\bf{84.3}$\\
			\hline
		\end{tabular}
		\vspace{-.3cm}
	\end{table*}

To systematically evaluate our method and study the contribution of each algorithm component, we provide an in-depth analysis of each individual design in our framework. We perform comprehensive ablative experiments on the W$32$ backbone to evaluate the capability of each component we claimed. All results are reported on the CrowdPose test dataset. The input image size is $512 \times 512$. As illustrated in Table \ref{tab1}, we present the the impact of each component.

In the first row of Table \ref{tab1}, we report the baseline results. The second row shows the the impact of the Global Relation Modeling (GRM). The third row shows results with the Multi-branch Feature Align (MFA). We can clearly see that each algorithm component is contributing significantly to the overall performance.

First, in order to solve the problem that previous bottom-up methods ignore to globally mine the relation in the feature learning process, we propose a Global Relation Modeling (GRM) module to efficiently capture the global relation among instances or environment. This strategy is used to combine features of the different stages to ensure a stronger discrimination for the current stage. Together with features of current stage, a dual-attention component is added to produce fused results. With this design, the current stage can take full advantage of prior information to extract more discriminative representations. As can be seen from Table \ref{tab1}, the proposed feature aggregation strategy increased baseline from  $72.7\operatorname{AR}$ to $73.2\operatorname{AR}$ by $0.5\operatorname{AR}$, proving the effectiveness of the strategy in dealing with the above problems.

Second, in order to solve the problem of receptive wild gap of different scale feature map fusion, a Multi-branch Feature Align (MFA) module is proposed to obatain refined local information. It has a high AP score ($66.8\operatorname{AP}$) owing to the deep connections and frequent feature aggregations inside the same level. This makes the low-level features sufficiently supervised resulting in satisfactory delicate spatial texture information, which benefits the keypoint localization.

Qualitative results are shown in Fig. \ref{fig3}. The first two rows are the visualization results of the COCO dataset, and the last two rows are the visualization results of the CrowdPose dataset. We compare our method with the state-of-the-art model \cite{ref19}, the first and third rows are the results of \cite{ref19}, and the second and fourth rows are the results of our method. It can be seen that in the case of adding our module, some occluded situations can also be solved very well, and the original wrong points can be correctly divided into groups. This proves that our proposed module effectively learns the global relation including some edge and texture information in the early stages of the network, and refines the representation of keypoint location feature.

 \renewcommand{\arraystretch}{1.15}
	\begin{table*}[t]
		\centering
		\caption{Comparisons on the COCO test-dev set. * means using refinement.}
\setlength{\tabcolsep}{8.24pt}
		\label{tab4}
		\footnotesize
		\begin{tabular}{l|c|lllll|lllll}
		    \hline
			Method & Input size &$\operatorname{AP}$ & $\operatorname{AP}^{50}$ & $\operatorname{AP}^{75}$ & $\operatorname{AP}^{M}$ & $\operatorname{AP}^{L}$ & $\operatorname{AR}$ & $\operatorname{AR}^{50}$ & $\operatorname{AR}^{75}$ & $\operatorname{AR}^{M}$ & $\operatorname{AR}^{L}$\\
			\hline
			\multicolumn{10}{c}{single-scale testing}\\
			\hline
            OpenPose* & $-$ & $61.8$ & $84.9$ & $67.5$ & $57.1$ & $68.2$ & $66.5$ & $-$ & $-$ & $-$ & $-$\\
            Hourglass~\cite{ref9} & $-$ & $65.5$ & $86.8$ & $72.3$ & $60.6$ & $72.6$ & $-$ & $-$ & $-$ & $-$ & $-$\\
			PifPaf & $-$ & $66.7$ & $-$ & $-$ & $62.4$ & $72.9$ & $-$ & $-$ & $-$ & $-$ & $-$\\
            SPM*~\cite{ref41} & $-$ & $66.9$ & $\bf{88.5}$ & $72.9$ & $62.6$ & $73.1$ & $-$ & $-$ & $-$ & $-$ & $-$\\
			HGG & $-$ & $\bf{67.6}$ & $85.1$ & $73.7$ & $\bf{62.7}$ & $74.6$ & $-$ & $-$ & $-$ & $-$ & $-$\\
            PersonLab & $1401$ & $66.5$ & $88.0$ & $72.6$ & $62.4$ & $72.3$ & $71.0$ & $-$ & $-$ & $\bf{66.1}$ & $77.7$\\
            HrHRNet-W$32$ & $512$ & $66.4$ & $87.5$ & $72.8$ & $61.2$ & $74.2$ & $-$ & $-$ & $-$ & $-$ & $-$\\
            HrHRNet-W$48$ & $640$ & $68.4$ & $88.2$ & $75.1$ & $64.4$ & $74.2$ & $-$ & $-$ & $-$ & $-$ & $-$\\
			\hline 
			Our approach (HRNet-W$32$) & $512$ & $67.6$ & $88.4$ & $\bf{74.5}$ & $62.1$ & $\bf{76.3}$ & $\bf{72.8}$ & $\bf{91.2}$ & $\bf{78.7}$ & $65.9$ & $\bf{82.1}$\\
			Our approach (HRNet-W$48$) & $640$ & $\bf{69.1}$ & $\bf{89.0}$ & $\bf{76.0}$ & $\bf{64.9}$ & $\bf{76.2}$ & $\bf{74.8}$ & $\bf{92.7}$ & $\bf{80.7}$ & $\bf{69.1}$ & $\bf{82.6}$\\
			\hline
			\multicolumn{10}{c}{multi-scale testing}\\
			\hline
			AE*~\cite{ref9} & $512$ & $65.5$ & $86.8$ & $72.3$ & $60.6$ & $72.6$ & $70.2$ & $-$ & $-$ & $64.6$ & $78.1$\\
            DirectPose~\cite{ref42} & $800$ & $64.8$ & $87.8$ & $71.1$ & $60.4$ & $71.5$ & $-$ & $-$ & $-$ & $-$ & $-$\\
			SimplePose~\cite{ref43} & $512$ & $68.1$ & $-$ & $-$ & $\bf{66.8}$ & $70.5$ & $72.1$ & $-$ & $-$ & $-$ & $-$\\
            HGG & $512$ & $67.6$ & $85.1$ & $73.7$ & $62.7$ & $74.6$ & $71.3$ & $-$ & $-$ & $-$ & $-$\\
			PersonLab & $1401$ & $68.7$ & $89.0$ & $75.4$ & $64.1$ & $75.5$ & $\bf{75.4}$ & $-$ & $-$ & $\bf{69/7}$ & $\bf{83.0}$\\
            Point-Set & $640$ & $68.7$ & $\bf{89.9}$ & $\bf{76.3}$ & $64.8$ & $75.3$ & $74.8$ & $-$ & $-$ & $69.6$ & $82.1$\\
            HrHRNet-W$32$ & $512$ & $-$ & $-$ & $-$ & $-$ & $-$ & $-$ & $-$ & $-$ & $-$ & $-$\\
            HrHRNet-W$48$ & $640$ & $\bf{70.5}$ & $89.3$ & $\bf{77.2}$ & $\bf{66.6}$ & $75.8$ & $-$ & $-$ & $-$ & $-$ & $-$\\
			\hline 
			Our approach (HRNet-W$32$) & $512$ & $\bf{68.9}$ & $88.5$ & $76.0$ & $64.3$ & $\bf{75.7}$ & $74.4$ & $\bf{92.3}$ & $\bf{80.7}$ & $68.9$ & $82.1$\\
			Our approach (HRNet-W$48$) & $640$ & $70.2$ & $88.9$ & $\bf{77.2}$ & $66.2$ & $\bf{76.3}$ & $\bf{76.1}$ & $\bf{93.1}$ & $\bf{82.2}$ & $\bf{70.9}$ & $\bf{83.3}$\\
			\hline
		\end{tabular}
		\vspace{-.3cm}
	\end{table*}

\begin{figure*}[t]
	\centering
	\includegraphics[height = 0.135\textwidth]{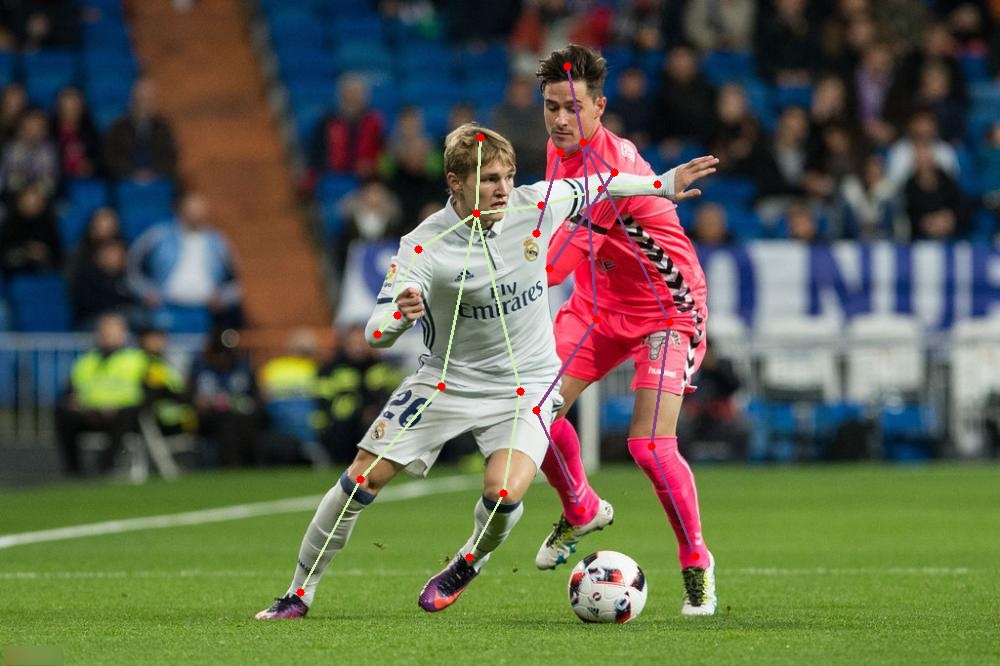}
	\includegraphics[height = 0.135\textwidth]{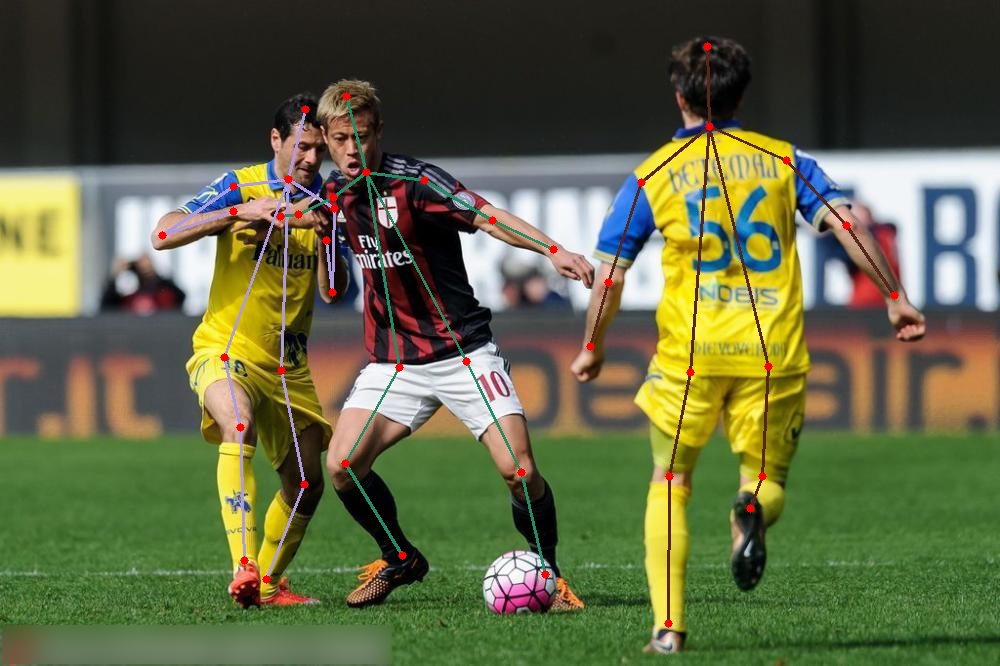}
    \includegraphics[height = 0.135\textwidth]{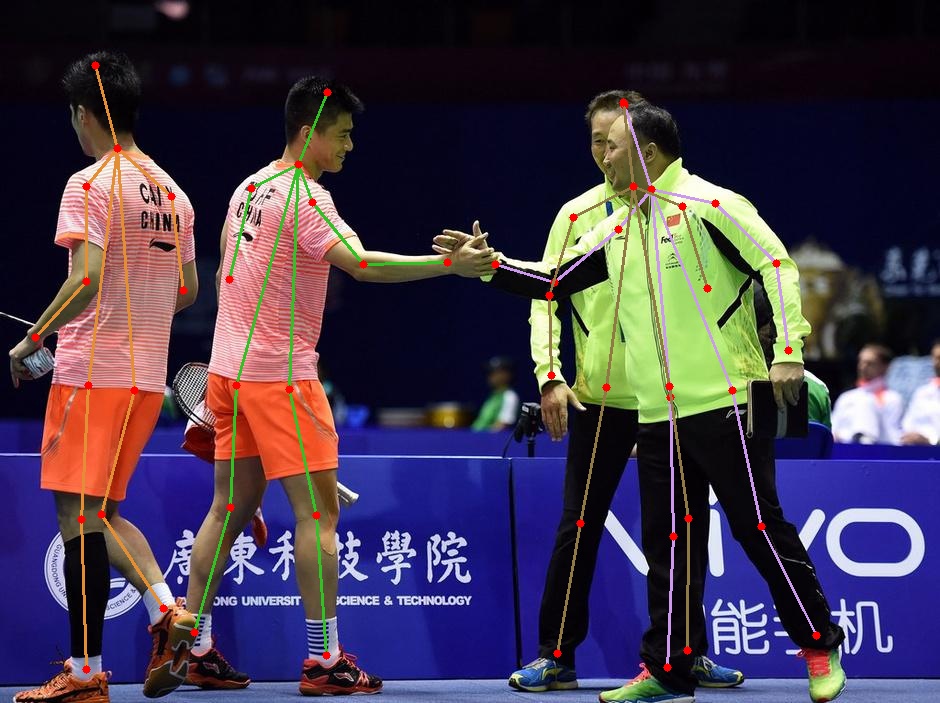}
    \includegraphics[height = 0.135\textwidth]{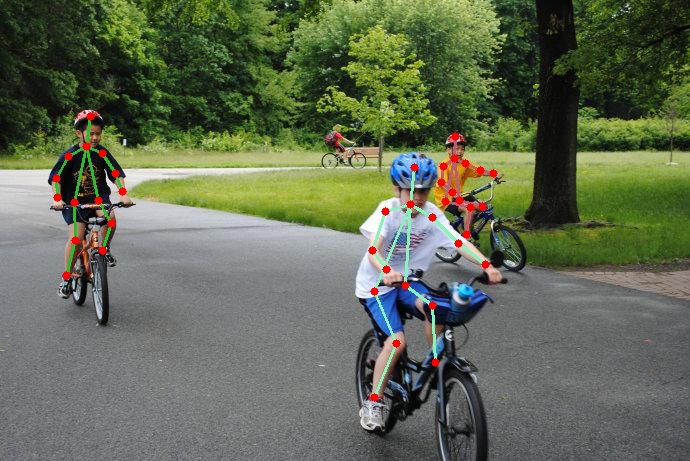}
    \includegraphics[height = 0.135\textwidth]{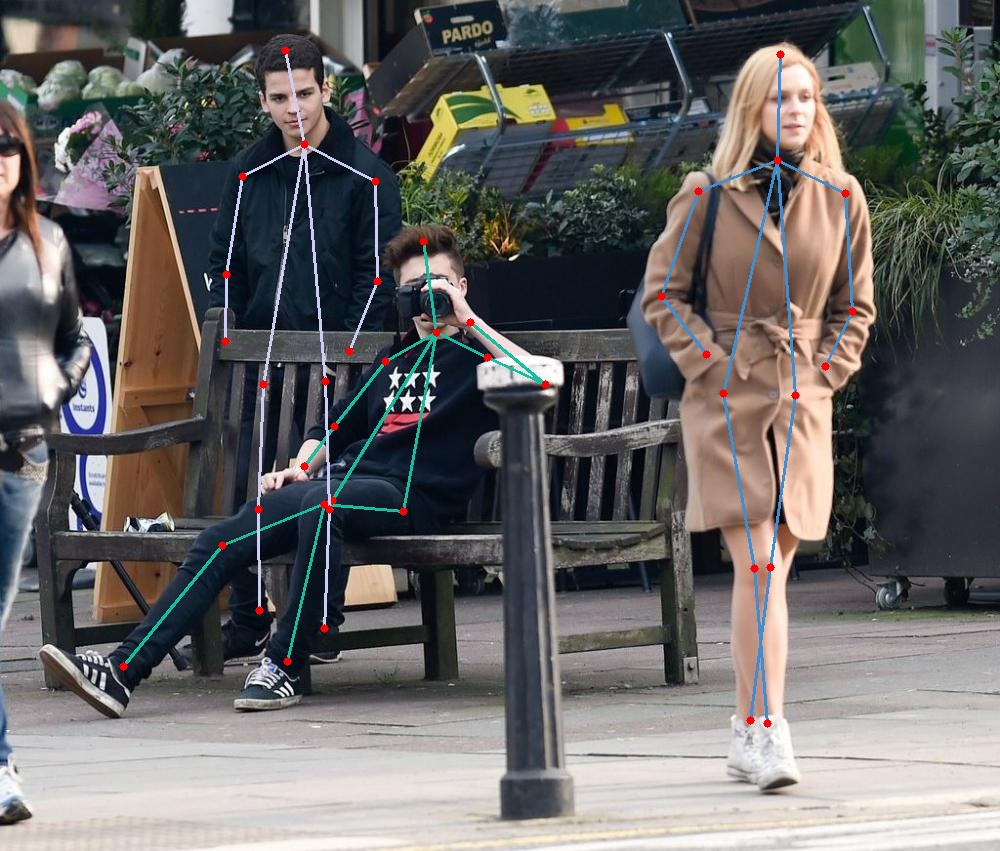}\\
	\includegraphics[height = 0.221\textwidth]{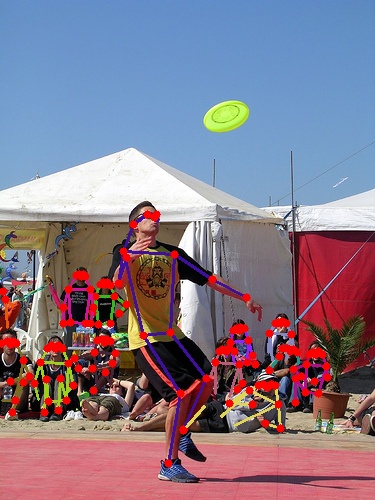}
	\includegraphics[height = 0.221\textwidth]{422.jpg}
    \includegraphics[height = 0.221\textwidth]{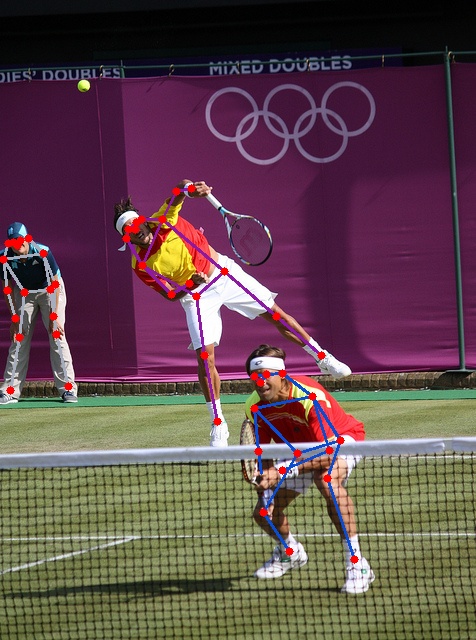}
    \includegraphics[height = 0.221\textwidth]{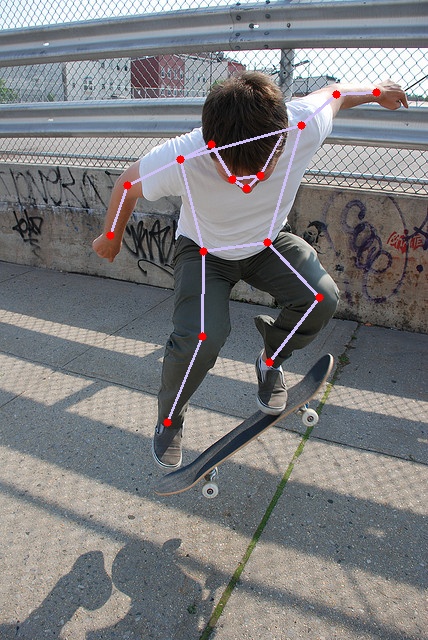}
    \includegraphics[height = 0.221\textwidth]{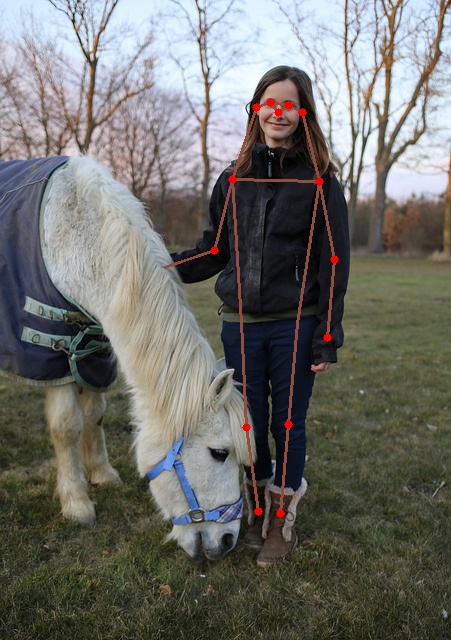}
    \includegraphics[height = 0.221\textwidth]{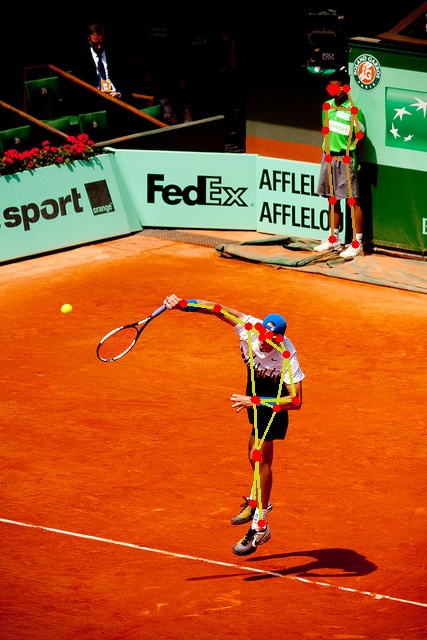}\\
	\caption{Qualitative results of some example images in the CrowdPose (top) and COCO (botttom) datasets: containing multiple persons, occlusion, and common imaging artifacts.}
    \label{fig4}
\end{figure*}

\subsubsection{Comparison with State-of-the-art Methods}\ 

{\bf{Results on CrowdedPose test set.}} In Table \ref{tab2}, we show the test-set results for our model trained on CrowdPose.  This dataset is focused on much more challenging images with severe occlusions. In this setting, our model shows its full potential and obtains good performance among all methods with  $66.8\operatorname{AP}$, where we improve upon state-of-the-art by $1.1$ and $1.2\operatorname{AP}$ points for single and multi-scale testing, respectively. Our approach with W$48$ as the backbone achieves $68.6\operatorname{AP}$ and is better than DEKR-W$48$ ($67.3 \operatorname{AP}$). With multi-scale testing, our approach with W$48$ achieves $69.1 \operatorname{AP}$ score, leading to $1.1 \operatorname{AP}$ gain over DEKR-W$48$. This proves that our model does benefit from being trained on a dataset in which occlusions are common, and results in better generalization to challenging images. Overall, we show that our method can outperform bottom-up approaches on difficult scenarios with severe occlusions, where reasoning about keypoint detection and grouping jointly has a clear benefit.

Qualitative results are shown in the top row of Fig. \ref{fig4}. We observe our model can deal with person overlapping ($1$st to $3$rd examples), person scale variations ($4$th example) and occlusion (the last example), showing the efficacy on producing robust pose estimation in various challenging scenes.

{\bf{Results on COCO val. set.}} 
Table \ref{tab3} summarizes the results on COCO$2017$ val-dev dataset. From the results, we can see that our method with only single scale test outperforms many other models. Our proposed network ($67.3 \operatorname{AP}$) outperforms HrHRNet by $0.2 \operatorname{AP}$. In addition, our model received a high AR score ($75.5 \operatorname{AR}$) in the COCO val dataset. If we further use multi-scale test, our model achieves $77.0 \operatorname{AR}$, outperforming all existing bottom-up methods by a large margin.

{\bf{Results on COCO test. set.}} 
Table \ref{tab4} shows the comparisons of our method and other state-of-the-art methods.  On COCO benchmark, compared to the HrHRNet \cite{ref18} with the same input size, our small and big networks receive $1.2$ and $0.7 \operatorname{AP}$ improvements, respectively. With multi-scale for testing, our model can obtain an $\operatorname{AP}$ of $70.2$.

Qualitative results on COCO dataset are shown in the bottom row of Fig. \ref{fig4}. We can see that our model is effective in challenging scenes, e.g., appearance variations and occlusion.


\section{Conclusion}
We propose a convolutional neural network with rich context for bottom-up human pose estimation. In our method, we design a light-weight backbone network, inheriting the multi-stage feature extraction process from HRNet, which aggregates features of different resolutions. In order to overcome some problems in previous bottom-up methods, we first propose a Global Relation Modeling (GRM) module to efficiently capture the global relation among instances or environment in the feature learning process. And then we propose a Multi-branch Feature Align (MFA) module, which aggregates feature from multiple branches to align fused feature and obtain refined local keypoint representations. The results show that the various indicators of COCO and CrowdPose data set have good performance, and prove the superior ability of our model to estimate posture in densely populated images.

\section*{Acknowledgments}
This work was supported partly by the National Natural Science Foundation of China (Grant No. 62173045, 61673192),
and partly by the Fundamental Research Funds for the Central
Universities (Grant No. 2020XD-A04-2).


\vspace{11pt}

\vspace{-33pt}
\begin{IEEEbiography}[{\includegraphics[width=1in,height=1.25in,clip,keepaspectratio]{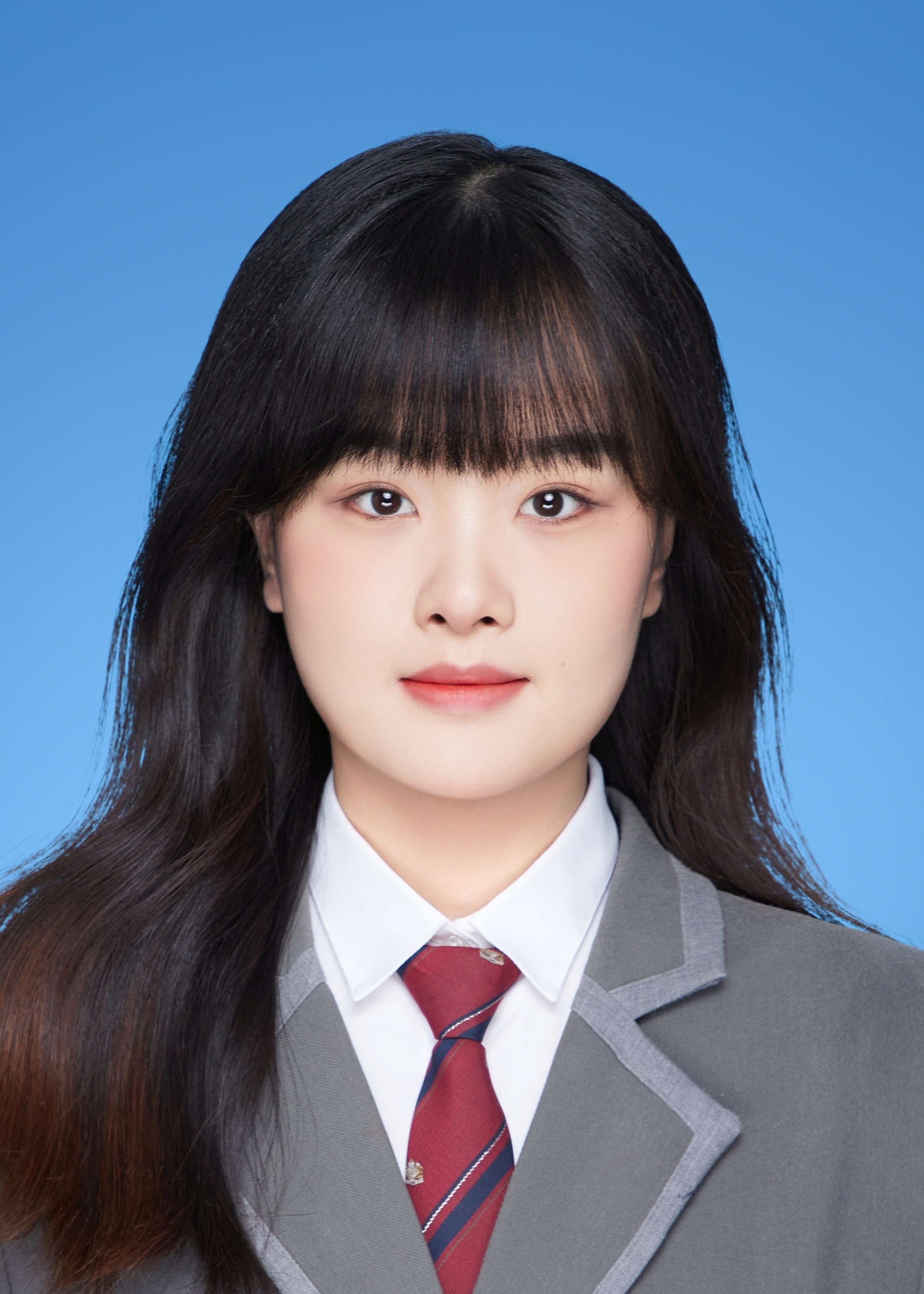}}]{Ruoqi Yin}
She currently is a bachelor in Artificial
Intelligence School, Beijing University of Posts and
Telecommunications, Beijing, China. Her research
interests include image processing, pose estimation, and deep learning.
\end{IEEEbiography}

\begin{IEEEbiography}
[{\includegraphics[width=1in,height=1.25in,clip,keepaspectratio]{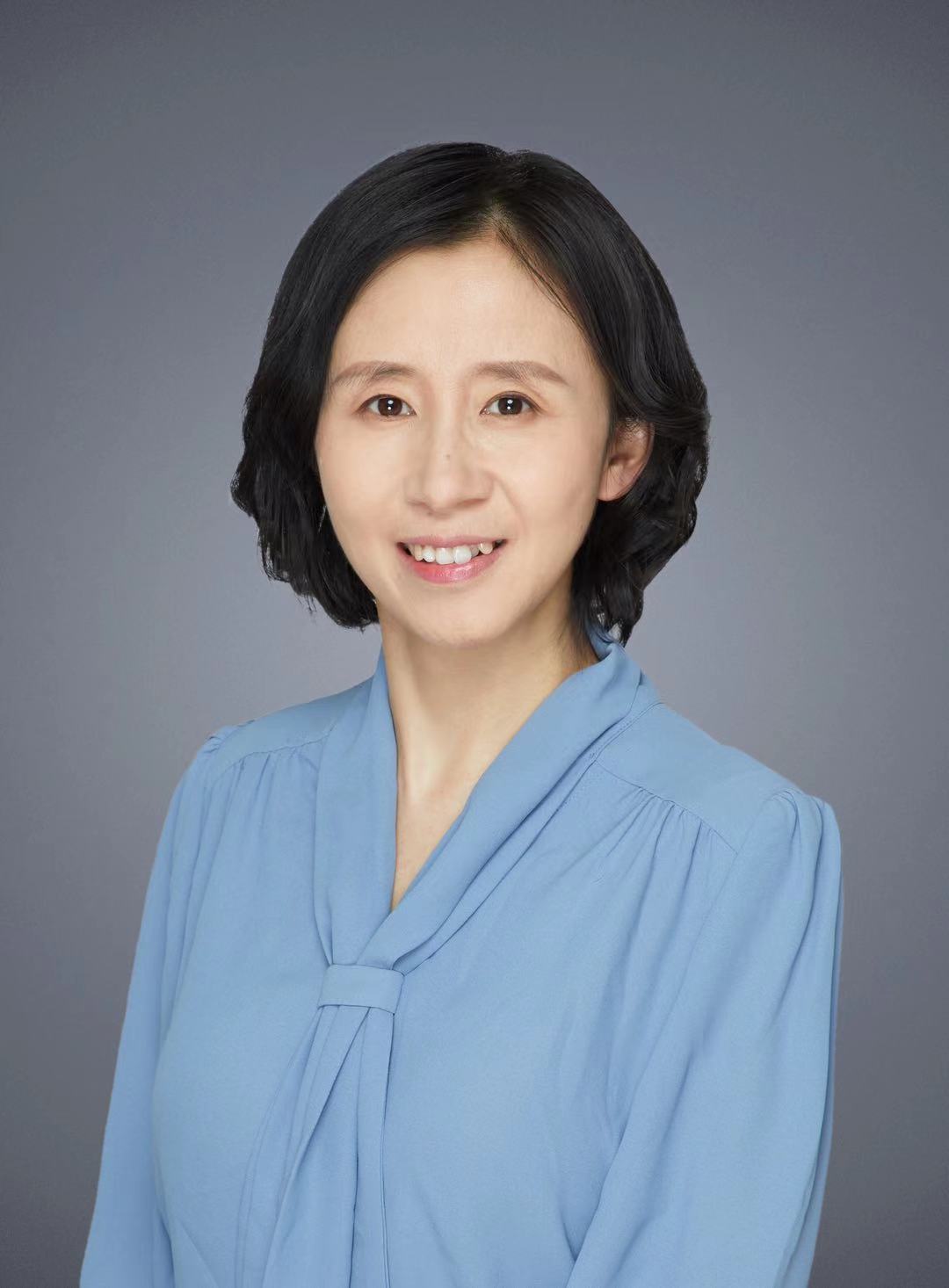}}]{Jianqin Yin}
(Member, IEEE) received the Ph.D. degree from Shandong University, Jinan, China, in 2013. She is currently a Professor with the School of Artificial Intelligence, Beijing University of Posts and Telecommunications, Beijing, China. Her research interests include service robot, pattern recognition, machine learning, and image
processing.
\end{IEEEbiography}

\vspace{11pt}

\vfill

\end{document}